# Machine Learning Approach to Polymerization Reaction Engineering: Determining Monomers Reactivity Ratios

Tung Nguyen and Mona Bavarian *

*Department of Chemical and Biomolecular Engineering, University of Nebraska-Lincoln*

*Lincoln, NE, 68588*

* Corresponding author: Mona Bavarian

Email: mona.bavarian@unl.edu

**Abstract**

In copolymerization, the monomers' reactivity ratios play an important role in shaping the final copolymer properties. Thus, the knowledge of reactivity ratio is essential to the polymerization reaction engineering. However, the experimental methods of determining reactivity ratios are laborious and require many repetitions and testing at different compositions. Therefore, computational methods for determining the reactivity ratios based on the chemical structures of the reactants have attracted significant attention. Here, we demonstrate how machine learning enables the prediction of comonomers reactivity ratios based on the molecular structure of monomers. We combined multi-task learning, multi-inputs, and one of the state-of-the-art machine learning model – Graph Attention Network – to build a model capable of predicting reactivity ratios based on the monomers' chemical structures. We found that the interpretable characteristics of Multi-Input-Multi-Output Graph Attention Network, along with its capability to learn chemical features facilitates the accurate prediction of the reactivity ratios. Such a predictive tool can be used in combination with macroscopic kinetics models to design new copolymers.

1.      Introduction

Copolymers are used in many different coating applications because their properties can be tailored by changing their constituent monomers.[1]–[3] The degree of incorporation of a monomer into a copolymer defines the final properties of these materials. Thus, understanding of copolymerization kinetics is essential to the materials design and synthesis.[4]–[6] In a copolymerization reaction, monomers' reactivity ratios have vital roles in the polymer chain microstructures, and they define the compositions and chain lengths of copolymers.[7] There have been many pioneering efforts to determine the reactivity ratios, such as the Mayo-Lewis method or the nested-iterative error-in-variables model (EVM).[8]–[10] However, most methods require a considerable number of experiments to allow for the accurate estimation of the reactivity ratios.[11], [12]

Recently, atomic-level simulation methods, such as Density Function Theory (DFT), have been applied to predict monomers' reactivity ratios.[13], [14] However, the computational cost of these calculations is quite significant, and several calculations with different levels of theory and basis sets are required. As an example, to determine the reactivity ratios, Johann et al.[13] first determined the activation energy of the transition state and reagents via DFT calculations and then used the transition state theory to estimate the rate constants ($k_p$) of the copolymer reactions. Finally, the reactivity ratio was determined by considering all possible propagation rate constants of the copolymer reactions.[13], [15] Taking DFT approach requires a significant number of calculations. Thus, due to the high computational cost of DFT methods, other computational tools capable of determining the monomers' reactivity ratios are of high interest.

In the past few years, the use of Machine Learning (ML) methods to predict the properties of materials proved promising in many fields, e.g., heterogenous catalyst and pharmaceutical research.[16], [17] ML methods have been also utilized in the polymer field to predict material properties, such as glass transition temperature, solvation free energy, and solubility.[18]–[22] Although many efforts have been devoted to predicting homopolymers' properties,[18], [21] the use of ML in predicting copolymers' properties is in its infancy. For instance, to date there is no report on the use of ML for predicting the reactivity ratios of copolymers. To study reaction kinetics, a reaction network for establishing the connections between monomers and different pathways to product is required. Several approaches have been proposed for developing ML models to study and predict properties of copolymers.[23]–[25] However, building ML models for copolymers has its own challenges associated with representation of copolymers' structures. For instance, for the prediction of copolymer using a graph representation, Aldeghi et al.[24] developed an ML model and included several crucial parameters, such as stoichiometry and degree of polymerization; nevertheless, their model could not distinguish between different isomeric monomers.[24] Because an isomeric structure is strongly associated with the reactivity ratio of monomers,[26], [27] Aldeghi's model is not suitable for predicting the reactivity ratios of monomers. Without suitable copolymer's representation, the encoding of sophisticated chemical structures is still an obstacle for ML model development.

Herein, we present an approach to utilize graph representation with multiple inputs to predict the reactivity ratios of monomers. We rely on model's learning based on monomers structures as well as the combined structure of two monomers representing a single unit of a copolymer. Our training dataset consists of more than 100s of different monomers' combinations collected from the experimental data. We used 13 different representations of atom and bond features to characterize and capture an atom and its local environment within the molecule; additional information of features are described elsewhere.[19] We then train the multi-task learning in the Graph Attention Network (GAN) model using these features to predict reactivity ratios of monomers. This work demonstrates that using an interpretable graph model with an attention mechanism and learning from multiple inputs result in improving the predictive performance of the ML model.

The organization of the rest of this paper is as follows. Section 2 presents the model development for copolymers. Different essential operations to build a graph attention model are described in this section. Section 3 describes the model performance in predicting the monomers reactivity ratios. Finally, Section 4 presents some concluding remarks.

## 2. Model Development

### 2.1. Dataset Preprocessing

Fig. 1 illustrates a workflow for predicting reactivity ratios of monomers in this study. A dataset of reactivity ratios of monomers was gathered from the polymer handbook.[28] As shown in Fig. 1, the dataset was first pre-processed by removing the monomers with no reported values for their reactivity ratios as well as the ones with untranslatable chemical structures. Then, the structure of remaining monomers was encoded in the Simplified Molecular-Input Line-Entry System (SMILES) chemical notation language [29] via an open-source cheminformatics toolkit, RDKit.[30] Next, reactivity ratios were processed. Originally, the dataset of reactivity ratios of monomers had a highly positive skewed distribution as shown in Fig. S1 A-B of Supplementary Data (SD). To mitigate the impact of imbalanced dataset on the model's performance,[31] we applied a square root transformation.[32] As shown in Fig. S1 C-D, the final distribution of reactivity ratios is shifted into an acceptable region with skew values under 0.5.[32] Then, the dataset is further standardized for our model development. Next, the dataset is shuffled and split with a split ratio of 70/10/20, where 70, 10, and 20% of the dataset are used for the training, validation, and testing set, respectively.

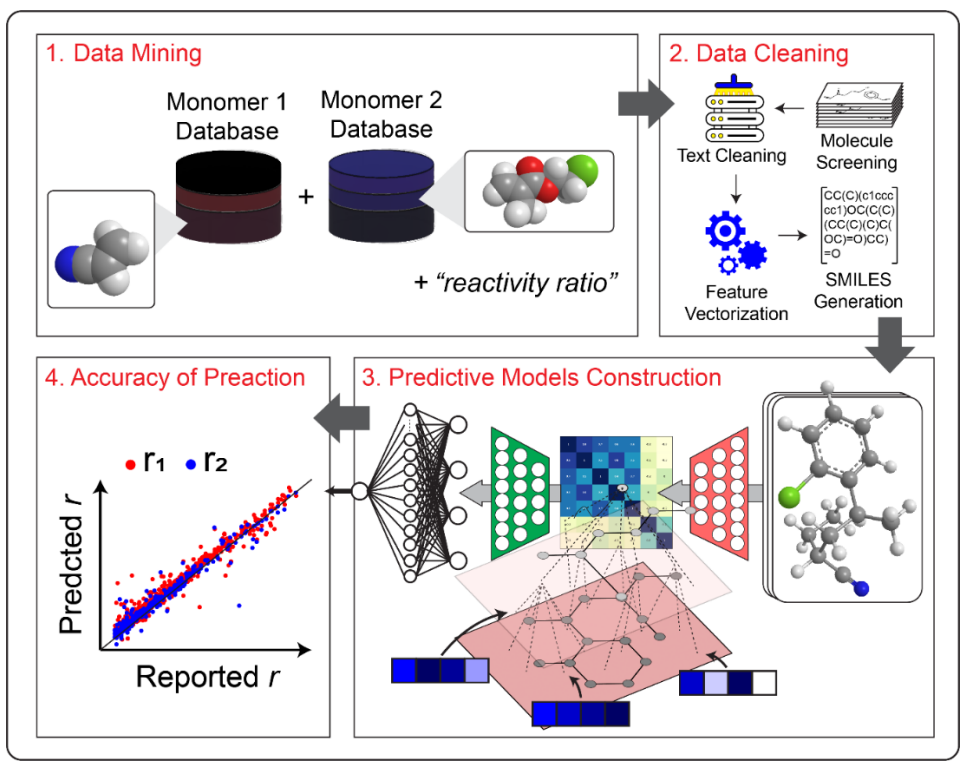

**Fig. 1.** Workflow of predicting reactivity ratios of monomers. (1) Data mining: Reactivity ratios for both monomers are collected; (2) Data cleaning & processing: monomers' SMILES are cleaned/preprocessed, and copolymers' SMILES are generated; (3) ML model construction: the ML model – Multi-Input-Multi-Output Graph Attention Network (MIMO GAN) – is developed, and the learning environment is evaluated via the Pearson Correlation Matrix (PCM); (4) Prediction result: monomers reactivity ratios are predicted.

## 2.2. SMILES Copolymers Preprocessing

SMILES is commonly used as a chemical line notation for translating chemistry knowledge into a machine understandable format, and thus it is successfully applied for predicting the properties of small single-unit chemicals.[33] Because many polymers consist of randomly distributed units, they do not have unique SMILES representations. Thus, there is no common agreement on representing macromolecules.[34] There are several successful pioneering efforts, such as the polymer genome,[35] to develop tools for predicting polymer properties from repeating-units. However, those developed tools are applicable to homopolymers. Recently, Lin et al.[36] proposed BigSMILES – a new structurally-based identifier to support the stochastic representation of repeating units of polymers by adding additional bonding descriptors. Further development is needed to use BigSMILES string in a graph model. In this study, we have applied several assumptions for model development. Similar to homopolymers that monomers SMILES used as the descriptor in many ML models[21], [33], [37], copolymers containing the SMILES of the two monomers are used in this work. By following a typical growth mechanism of a copolymer,[38] we generate a copolymer dataset of ~ 4000 copolymers. Table S1 of SD shows an example dataset of several SMILES structures of monomers and copolymer used in this study.

## 2.3. Methodology and Computational Details

### 2.3.1. Graph Attention Network (GAN)

A graph structure for a molecule in its basic form represents the atoms that make the material and describes the bonding types. Hence, it is necessary to use a ML model that has the capability to learn the molecules' structures from the corresponding graph structures.[39] Since its first application in 2009,[40] the graph neural network (GNN) has been utilized in many studies to predict properties of small molecules and polymers.[19], [41], [42] However, the models were limited to single repeating-unit homopolymers. Furthermore, the lack of distinguishable nodes' embeddings on a computational graph,[43] to embed a long copolymer's structure, impedes the GNN's power in predicting copolymers' properties. As a result, the GNN model cannot differentiate target nodes from the neighboring nodes, failing to provide reasonable predictions.[44] To improve the decision-making of GNN, an attention mechanism is utilized.[45] Here, an attention mechanism assigns different weights to different nodes so that the model learns more information about its neighbors and local environment. In this study, we applied the Graph Attention Networks (GAN) with the Attentive Fingerprint (Attentive FP), proposed by Xiong et al.,[19] to predict the reactivity ratios of monomers. The model relies on both the recursive neural network (RNN) workflow and an attention mechanism; they are employed to focus on the most important features of the input by considering its neighbors and following a self-attention strategy. The attention mechanism is built based on three methods. The first method is building three essential operations for an attention mechanism – alignment, weighting, and context. Equation 1, 2, and 3 are used for alignment, weighting, and context operations, respectively.

$$e_{vu} = leaky\_relu(W \cdot [h_v, h_u]) \quad (1)$$

$$a_{vu} = softmax(e_{vu}) = \exp(e_{vu}) / \sum_{u \in N(v)} \exp(e_{vu}) \quad (2)$$

$$C_v = elu\left(\sum_{u \in N(v)} a_{vu} \cdot W \cdot h_u\right) \quad (3)$$

Where $v$ is a target node, which is a specific atom, $N(v)$ represents all neighbors of node $v$, $h_v$ is a state vector of node $v$, $h_u$ is a state vector of neighbor atom (node $u$), and $W$ is a learnable weight matrix, which indicates a relationship between the target node and its neighbors. After obtaining the alignment scores vector from the $alignment$ (Eq. 1), we apply $softmax$ on this vector from the $weighting$ step (Eq. 2) to obtain the attention weights. By using $softmax$, all the values in the vector are normalized. Then, the context vector is generated via $elu$ function, which allows a non-zero slope for the negative part of the ReLU function.[46] When the feature score is close to 1, it has more influence on the decoder output; however, the feature score is less important when its value is close to 0. Lastly, the feature is nullified as its score value is negative, and hence the model ignores this feature. These operations help the model to obtain the context of a target atom by focusing on its neighbors and local environment.

The second method is building a feasible model. In this model, a gate recurrent unit (GRU) is utilized so that it can retain and filter information from the past and current step. The GRU is similar to RNN, but two additional reset and update gates are added in GRU in order to help the model filter out non-relevant information during the iteration process. Hence, GRU can improve the memory capacity of RNN. This action improves the learning performance and allows for distinguishing different atoms. In the model, the GRU is implemented in a messaging and readout phase, and their mathematical formulations are shown in Equations 4 and 5, respectively.

$$C_v^{k-1} = \sum_{u \in N(v)} M^{k-1}\left(h_v^{k-1}, h_u^{k-1}\right) \quad (4)$$

$$h_v^k = GRU^{k-1}(C_v^{k-1}, h_v^{k-1}) \tag{5}$$

Where $M^{k-1}$ is the message function at $k-1$ iteration. After the node features are learned, the nodes representation of these molecules are aggregated at the messaging phase. Here, the network uses the message function, $M^{k-1}$ to aggregate information from the neighbor nodes located on a graph for each target node. While the graph attention mechanism gathers information from all the neighbor nodes in the messaging phase to update the state in the read-out phase, the GRU takes an input from a previous state vector ($h_v^{k-1}$) of the target node and the attention context ($C_v^{k-1}$) from the neighbors. During the readout phase, the network uses the GRU function to update the current hidden state of the target node by using the information from the message phase and the previous hidden state of the target node.

The learned target nodes' representations are used to predict molecular properties through the read-out phase.[47] Further details on how these operations (Eq. 1-5) work and are utilized in GAN can be found elsewhere.[19] Finally, the model uses atom symbol, neighboring atoms, atom mask, bond types, and neighboring bond to distinguish the target node from its neighboring nodes; more details of each feature used in GAN are discussed elsewhere.[19] Here, an atom mask refers to the masking mechanism of atoms that do not carry useful information.[48]

### 2.3.2. Multi-Input-Multi-Output Graph Attention Network (MIMO GAN).

Since both monomers and copolymers influence the reactivity ratios of monomers, we used the SMILES of both two monomers and the corresponding copolymer. We also considered the benefit of the multi-task learning model as it tends to compromise the performance of learning tasks and improve the learning capabilities even with a sparse dataset;[48] hence the multi-task model to predict the reactivity of both monomers was approached. The network architecture used in this study is similar to Attentive FP.[19] Instead of using a single SMILES input, three inputs of monomers and copolymer were fed to the model.

### 2.3.3. Training and Evaluation Metrics

The main hyperparameters helping to improve the learning and prediction performance of a model are the fingerprint dimension, the layer weight regularizer, the learning rate, and the dropout rate. In this study, these noted features are utilized as hyperparameters; more details of hyperparameters can be found in Table S2. For the training protocol, we use the gradient descent with a batch size of 250 for training efficiency, the Adam optimizer,[49] and the initial learning rate of 5.4E$^{-3}$. The learning rate is scheduled to fall when performance stagnates after 10-20 epochs. At that stage, the lowest learning rate is set at ~1E$^{-6}$. The training is determined by a standard mean-square error (MSE) loss function, which computes the average of the squared distance between actual values and predicted values, shown in Equation 6, and the mean value over the last dimension is returned for the next evaluation step.

$$loss_i^j = \frac{1}{N}\sum_{i=1}^{N}\left(y_{pred,i}^j - y_{actual,i}^j\right)^2 \tag{6}$$

Where $y_{pred}$ and $y_{actual}$ is a predicted and reported reactivity ratio value, $N$ is a total number of data points, $loss_i$ is the loss of each target value, and $i$ is a number of data point; $j$ is the monomer's type index, (in this study, $j \in \{1,2\}$, representing for reactivity of monomer 1 and 2, respectively). To quantify the performance of MIMO GAN, the root-mean-square error (RMSE) and R-square ($R^2$) values are examined. For the development and computing of MIMO GAN, Python language and PyTorch library were utilized. The model was trained using NVIDIA Tesla V100 GPU with 32 GB of RAM with a single-core GPU.

## 3. Results & Discussion

### 3.1. Using the Interpretation Method of MIMO GAN for the Reactivity Ratio Predictions

#### 3.1.1. Evaluating interpretability of MIMO GAN

In general, ML models are used as black box,[50] and hence it is difficult for one to understand how the models use the learned features to predict the target values. Thus, it is crucial to understand the model's decision-making to assess the predicting capability of the model. To check whether the learned chemical features of MIMO GAN are interpretable, we evaluate a feature interpretation of MIMO GAN using SHapley Additive exPlanations (SHAP).[51] We found that an interpretable MIMO GAN extracted important node features to predict reactivity ratios of monomers; more information of node features can be found elsewhere.[19] This helps us to develop chemical insights in prediction of reactivity ratios of monomers. Fig. 2A illustrates an impact of eight important node features that are strongly associated with predictions of reactivity ratios of monomers. As can be seen, a high feature value, which is located in the positive SHAP values region, indicates a positive correlation to the model output; however, the higher feature value, which is located in the negative SHAP region, has a negative correlation to the output.

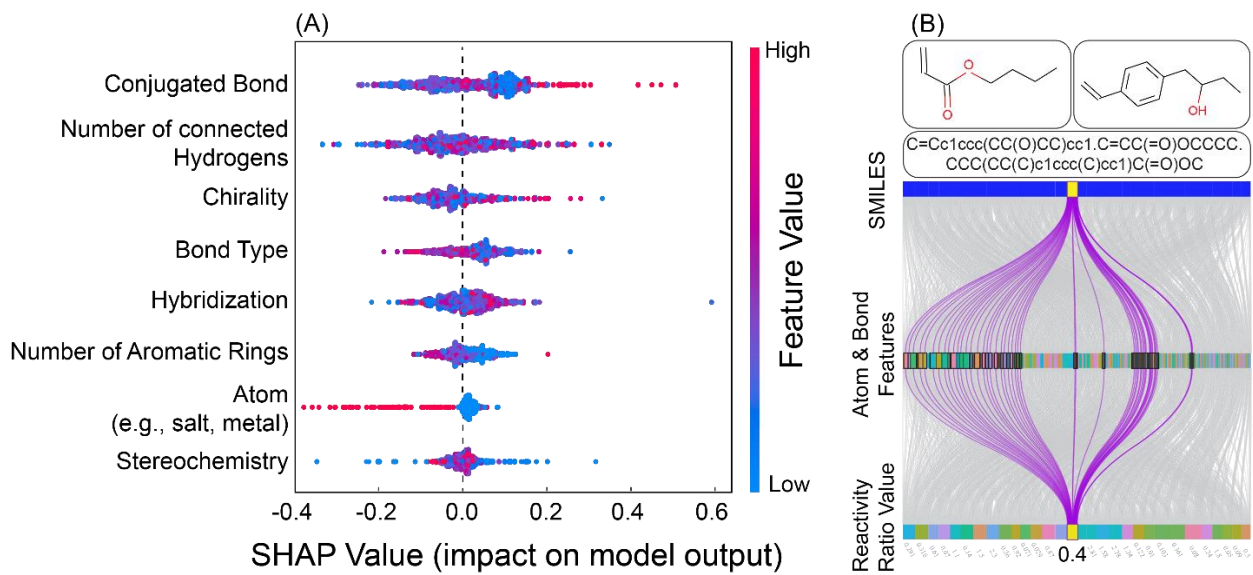

**Fig. 2.** (A) Summary plot of top eight node features on the reactivity ratios of monomers that impact on the reactivity ratio prediction. The atom & bond features are ranked based on an average score of their influences to the reactivity ratio prediction of trained samples. Each dot represents the impact of the corresponding pair of monomers and copolymer in the training set. Red and blue color indicates the high and low impact on the reactivity ratios, respectively. A positive SHAP value indicates a positive influence on the prediction, and a negative SHAP value indicates a negative influence on the prediction. (B) Illustration of how MIMO GAN interprets the learned features to predict the reactivity ratio values of monomers; 30 pairs of monomers and copolymers in the dataset are selected for a representation.

The first and most important node feature is the conjugated bond of a molecule. The conjugated bonds influence the stability and activation energy formation of monomers (i.e., how each monomer is formed when a radical approaches), affecting the reactivity ratios values of monomers.[52] Similarly, as the hydrogen transfer (directly correlates to a number of connected hydrogens) contributes to the formation of a final copolymer such that a final product is feasible,[53] this feature is another crucial feature that can be related to the reactivity ratios outcome, and hence MIMO GAN notes to pay attention to this feature. In contrast, the existence of special atoms (e.g., salt, metal complex) tends to have a greater negative impact on the reactivity ratios of monomers. This suggests that the reactivity ratios of monomers reduces and thus locates in a low value region due to different polarities from a contribution of solvents and monomers.[52], [54] To support our observations, several molecules with the corresponding reactivity ratios are demonstrated in Fig. S2, SD. By understanding and being able to interpret chemical features, MIMO GAN is able to interpolate/predict the reactivity ratios of monomers correctly. As illustrated in Fig. 2B, after the learned node features of the given molecules are aggregated and passed through a $softmax$ layer, MIMO GAN interprets features that are more important based on its chemical knowledge learning; then, the model extracts those features from the compounds to interpolate the reactivity ratios values. Overall, an analysis shows that the learned features from MIMO GAN can be interpretable so that the model can use them to predict reactivity ratios of monomers correctly.

### 3.1.2. Learning Chemical Features via Atom Similarity

Here, we investigate the relationship between different atom embeddings during the learning process by obtaining the Pearson correlation coefficient to determine the similarity coefficient between atom pairs. Fig. 3 illustrates the atom similarity matrix for a typical copolymer used in this study. As can be seen, every shaded square of each molecule has its corresponding Pearson's value noting the similarity of the target atom and the neighboring one; the Pearson's value closer to one indicates the higher atom similarity. As an example, for Octyl Acrylate (OA) monomer, after training, MIMO GAN learns that while there is 60% similarity between two atoms of C1 and C0, C1 shares ~40, 80, and 40% similarity with atom C2, C5-C11, and C12, respectively. A difference in the similarity is due to the difference in the electron densities, the number of attached hydrogen (H) atoms, and the distance between two atoms allocated to different functional groups. This is aligned with our expectation as reported in the literature.[55], [56] A similar appearance of the atom similarity is observed when atom C1 of OA becomes atom C11 of the copolymer shown in Fig. 3C.

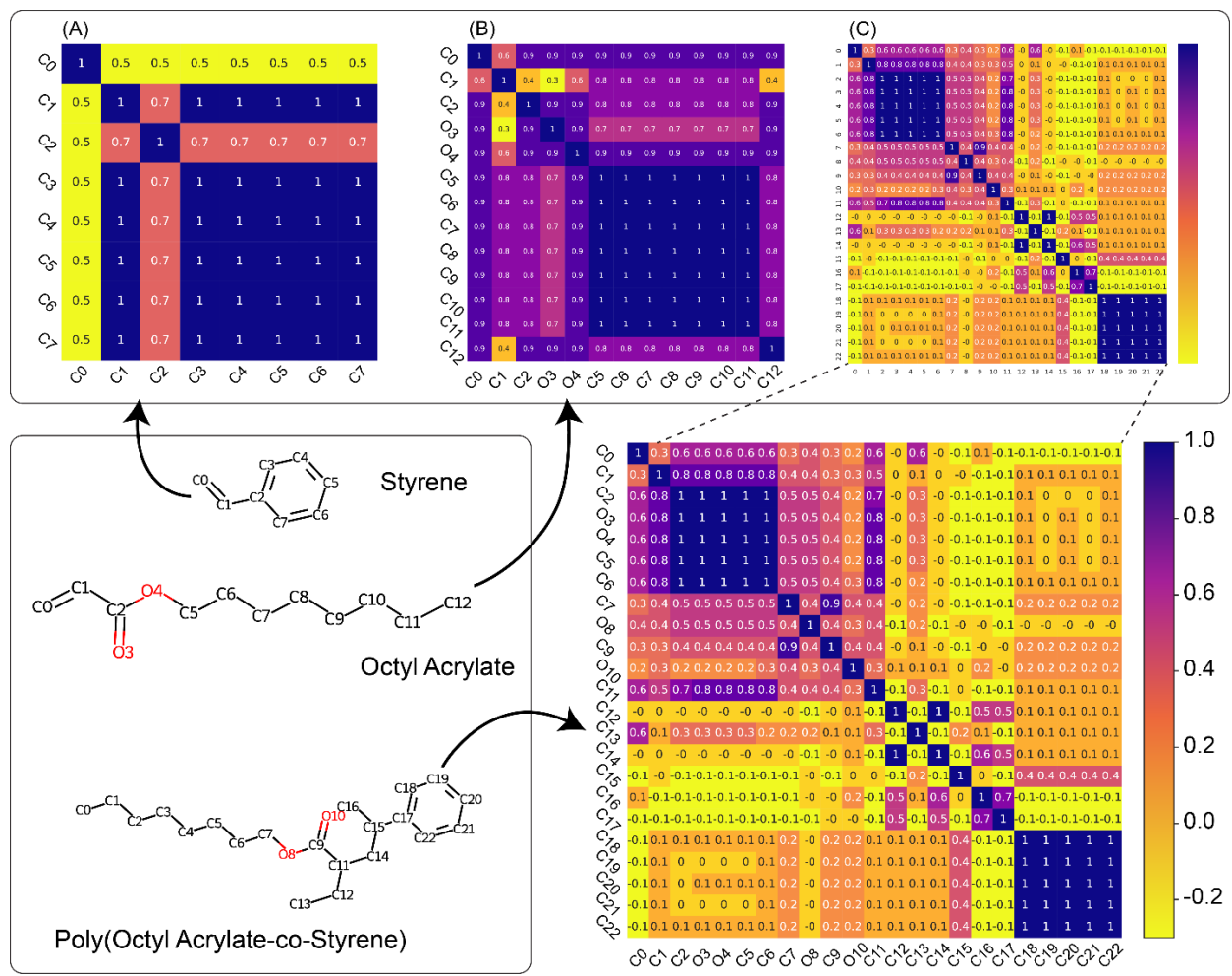

**Fig. 3.** Pearson correlation coefficient for Poly(Octyl Acrylate-co-Styrene) copolymer: (A) Styrene, (B) Octyl Acrylate, (C) Poly(Octyl Acrylate-co-Styrene). The similarity scores are shown inside the squares with the corresponding colors of each molecule. Blue patches indicates a positive correlation, yellow batches indicates a negative correlation, and Orange indicates neutral (or no correlation). The MIMO GAN is able to learn and correctly distinguish all subgroups of each molecule by separating them to different clusters.

Furthermore, we investigate different functional groups of each compound by MIMO GAN. The functional groups of each molecule are discernable and extracted into different clusters, highlighted by different color blocks; a darker color shows how close the neighboring nodes are to the target node, and a light color shows no correlation between target and neighboring nodes. This shows a strong agreement with the chemical intuition related to that molecule's structure. Overall, both the atom similarity visualization and its accurate functional groups extraction suggest that MIMO GAN can distinguish different functional groups and acquire a correct chemical knowledge of each molecule.

### 3.2. Monomers Reactivity Ratios Prediction

We assess the performance of the model by analyzing the parity plot of predicted reactivity ratios as illustrated in Fig. 4. As can be seen, by using multiple inputs and multi-task learning, many copolymers are accurately predicted, and we achieve the prediction accuracy up to ~ 83 ± 4.7 % and ~ 85 ± 5.2 % for $r_1$ and $r_2$, respectively. The results indicate that the MIMO GAN is able to capture information of chemical knowledge of the polymer reaction to interpret features properly and perform a good prediction of reactivity ratios. As all inputs contribute to the final latent representation, MIMO GAN borrows information from monomers to learn and improve the prediction capability. Thus, data agglomeration across different sources improves the accuracy of reactivity ratios prediction.

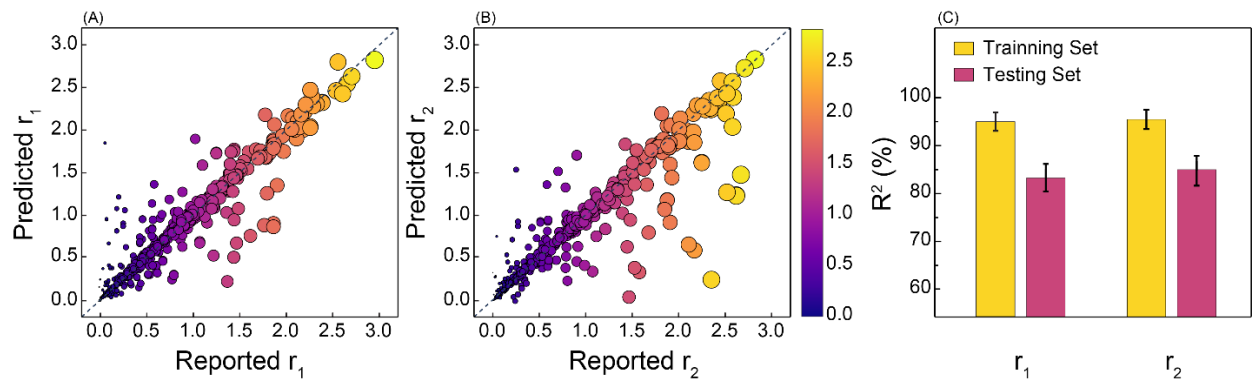

**Fig. 4.** Parity plot of reactivity ratios of monomers collected from the testing set: (A) reactivity of monomer 1; (B) reactivity of monomer 2. Both the size and the color of each dot represents the corresponding reported reactivity ratio value. (C) Bar chart represents $R^2$ value of the training and testing set of both reactivity ratios. Data are represented with an average of eight independent runs.

Despite the great interpretation capabilities of MIMO GAN, there are still limitations in the model such that the reactivity ratios of some of the compounds are incorrectly predicted. One of the major drawbacks of a graph model is a large amount of required computation and memory to store patterns, leading to an inefficient process on large graphs.[57] When MIMO GAN learns the fingerprints of copolymers, the model tends to lose information of nodes embeddings and hence cannot remember important features of copolymer structures to provide an appropriate predicted reactivity ratios value. Even though it is beneficial to use information from smaller graphs (i.e., monomers in this study), if the chemical features of monomers cannot be correctly acquired, then there is a conflict in learning substructures between monomers and copolymer. Without having distinguishable subgroups of the compounds during learning, MIMO GAN cannot interpret the chemical features of new molecules properly. Here, we look at Poly(Octyl Acrylate-co-Styrene), Poly(OA-S), (Fig. S3, SD); other copolymer pairs can be evaluated using the same approach. To understand errors attributed to Poly(OA-S), we assess the atom similarity and clusters extraction of MIMO GAN via the Pearson correlation coefficient of each compound (Fig. S4, SD). We decided to focus our analysis on Octyl Acrylate; the same method can also be applied to other compounds. As shown in Fig. S4E of SD, the atoms weights are properly assigned, and several substructures are discernable. However, this is not observed in Fig. S4B, SD and would result into inaccurate prediction of reactivity ratios for Poly(Octyl Acrylate-co-Styrene).

## 4. Conclusion

We demonstrated that an interpretable MIMO GAN has a potential of predicting the reactivity ratio values of copolymers without considering reaction conditions to some extent. The prediction results show that the MIMO GAN can capture the chemical knowledge of the copolymerization reactions to make a good prediction despite limitations in the representation of the sophisticated polymer structures. In addition, the attention mechanism at atomic and molecular level helps MIMO GAN to learn the chemical features of a given copolymer at the local and nonlocal environment, and hence it allows MIMO GAN to correctly capture the substructure patterns of the copolymer compound. Extracting hidden layers and attention weights helps the end-user to understand MIMO GAN's interpretable characteristics and gain more insight of MIMO GAN's decision-making for predicting the reactivity ratios of copolymers. Finally, the proposed workflow signifies the MIMO GAN's capability for predicting multiple properties of copolymers.

## 5. Author Information

**Corresponding Author**

Mona Bavarian – *Department of Chemical Engineering, University of Nebraska-Lincoln, Lincoln, Nebraska, 68588; orcid:* https://orcid.org/0000-0001-7689-773X Email: mona.bavarian@unl.edu

**Authors**

Tung Nguyen – *Department of Chemical Engineering, University of Nebraska-Lincoln, Lincoln, Nebraska, 68588;* orcid: https://orcid.org/0000-0003-2897-4889.

**Notes**

The authors declare no competing financial interests

**Acknowledgement**

This work was completed utilizing the Holland Computing Center of the University of Nebraska, which receives support from the Nebraska Research Initiative. This work was supported in part by the University of Nebraska-Lincoln through a Layman Award. T.N. also thanks Dr. Syed I.G.P Mohamed for the help on checking and cleaning some untranslatable structures.

**Appendix A. Supplementary data**

Supplementary data to this article can be found at

# Supplementary Data

**Machine Learning Approach to Polymerization Reaction Engineering: Determining Monomers Reactivity Ratios**

Tung Nguyen and Mona Bavarian *

*Department of Chemical and Biomolecular Engineering, University of Nebraska-Lincoln*

*Lincoln, NE, 68588*

\* Corresponding author: Mona Bavarian

Email: mona.bavarian@unl.edu

## Contents

**Main Section**

**Table S1.** Example of SMILES dataset used in this study.

| Monomer 1 | Monomer 2 | Copolymer | $r_1$ | $r_2$ |
|---|---|---|---|---|
| C=CC(=O)OC1CC(C)CCC1C(C)C | C=Cc1ccncc1 | CCC(CC(C)c1ccncc1)C(=O)OC1CC(C)CCC1C(C)C | 0.29 | 2.32 |
| C/C=C\C(=O)OCOC(F)(F)C(C)(F)F | C=C(C)C(=O)OC | COC(=O)C(C)(C)CC(C(=O)OCOC(F)(F)C(C)(F)F)C(C)C | 0.61 | 1.43 |
| C=Cc1ccc(CCCCCCCOS(N)(=O)=O)cc1 | C=C(C)C(=O)OCCCC | CCCCOC(=O)C(C)(CC)CC(C)c1ccc(CCCCCCCOS(N)(=O)=O)cc1 | 1.5 | 0.7 |
| Cc1cccc(Cl)c1N1C(=O)C=CC1=O | C=C(C)C(=O)OC | COC(=O)C(C)(C)CC1C(=O)N(c2c(C)cccc2Cl)C(=O)C1C | 0.029 | 0.56 |
| C=C1CC(C)N(Cc2cccc2)C1=O | C=C(C)C(=O)OC | CCC1(CC(C)(C)C(=O)OC)CC(C)N(Cc2ccccc2)C1=O | 2.28 | 0.38 |
| C=C(C)C(=O)OCCOC(C)=O | C=C(C)C(=O)OC | CCC(C)(CC(C)(C)C(=O)OC)C(=O)OCCOC(C)=O | 1.12 | 0.94 |
| C=C(C)C(=O)OC | C=C(C#N)C#N | CCC(C)(CC(C)(C#N)C#N)C(=O)OC | 0.057 | 0.03 |

**Table S2.** Summary of Hyperparameters used in this study.

| Name | Value |
|---|---|
| Batch size | 250 |
| Drop out | 0.05 |
| Epoch | 300 |
| Drop out | (0.03,0.44) |
| Weight decay | $1E^{-4}$ |
| Learning rate | $5.4E^{-3}$ |
| Minimum learning rate | $1E^{-6}$ |
| Gamma | 0.8 |
| Patience | 13 |
| Fingerprint dimension | 300 |
| Radius | 3 |
| T | 3 |
| Optimizer | Adam algorithm |
| Scale Type | RobustScaler (sklearn) |
| Quantile range (for scaling) | [5,95] |
| Split ratio | 7/1/2 (Train/Validation/Test) |

**Table S3.** Reactivity ratios comparison between the training, evaluation, and reported data of Poly[octyl acrylate-co-styrene].

| Training mode | | Evaluation mode | | Reported data | |
|---|---|---|---|---|---|
| $r_1$ | $r_2$ | $r_1$ | $r_2$ | $r_1$ | $r_2$ |
| 0.011 | 0.36 | 0.18 | 0.51 | 0.01 | 0.39 |

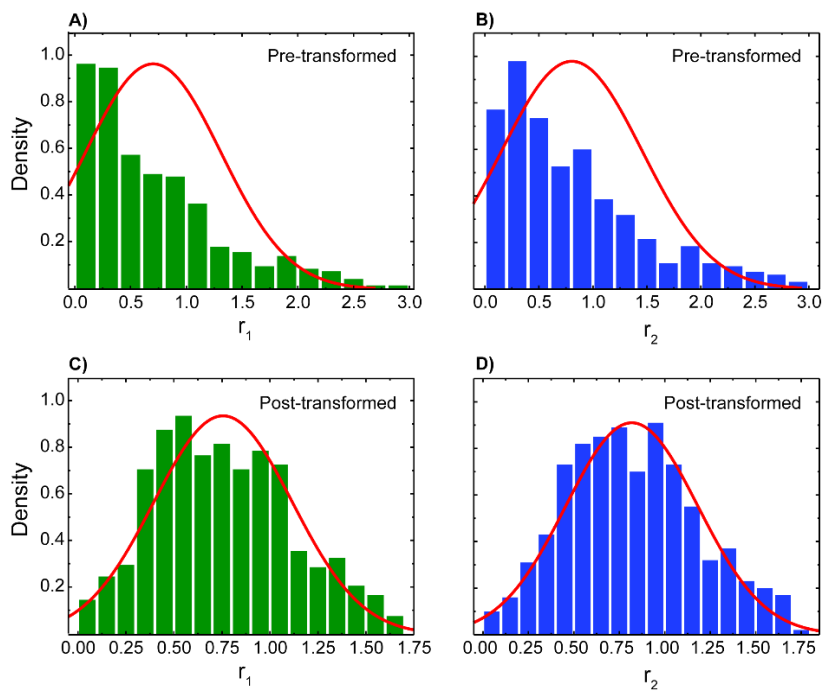

**Fig. S1.** Distribution of Reactivity Ratios. (A) original distribution of reactivity ratio value of monomer 1; (B) original distribution of reactivity ratio value of monomer 2; (C) post-processed distribution of reactivity ratio value of monomer 1 with the square root transformation; (D) post-processed distribution of reactivity ratio value of monomer 2 with the square root transformation.

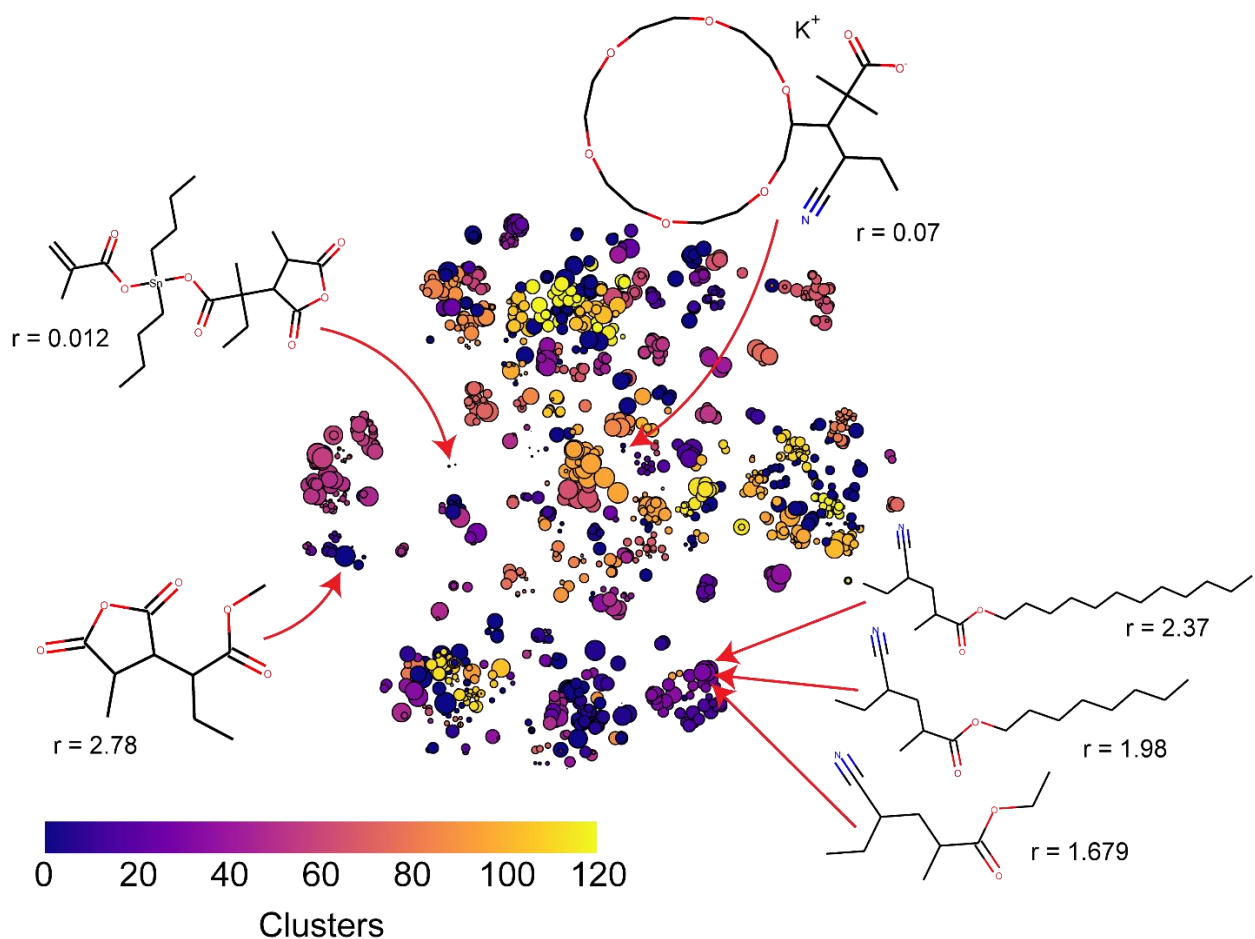

**Fig. S2.** Reactivity Ratios Chemical Space Visualization via t-SNE. Markers' size and color are representative of reactivity ratios and number of clusters, respectively. Several copolymers with the corresponding reactivity values are shown. Similar functional groups are identified as a cluster.

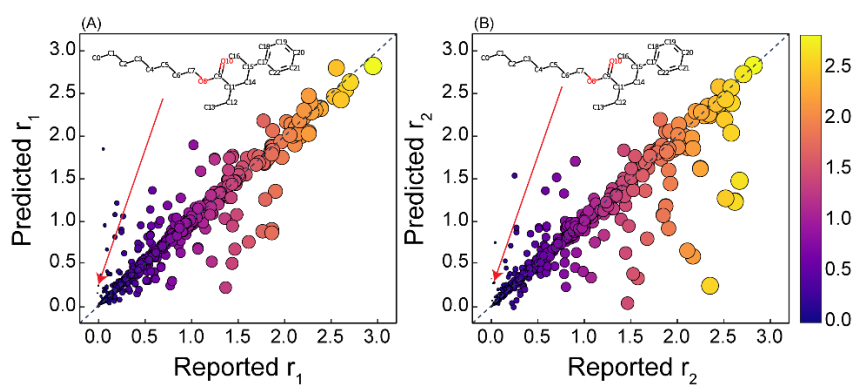

**Fig. S3.** Parity plot of reactivity ratios of monomers collected from the testing set: (A) reactivity ratio of monomer 1; (B) reactivity ratio of monomer 2. Both the size and the color of each dot represents the corresponding reported reactivity value. Poly[octyl acrylate-co-styrene] is selected for an analysis.

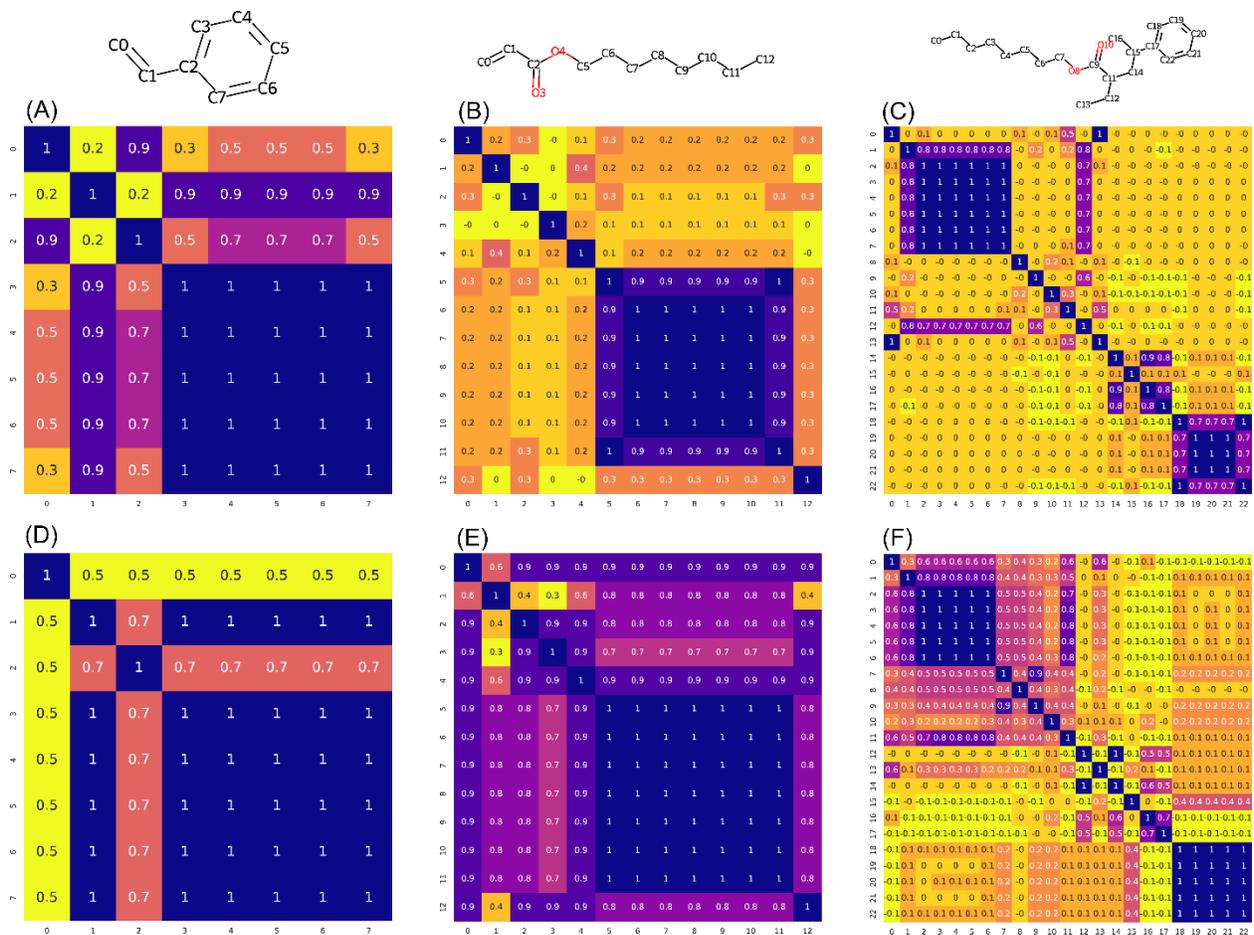

**Fig. S4.** Pearson correlation coefficient for Poly(Octyl Acrylate-co-Styrene), Poly(OA-S): (A,D) styrene (monomer 1); (B,E) octyl acrylate; (C,F) Poly(Octyl Acrylate-co-Styrene). The structures were evaluated in: (A-C) the evaluation mode; (D-F) the training mode. The predictions of both cases are acquired from MIMO GAN with similar selected hyperparameters. All figures are retrieved from radius of 3.